# Semiconductor Wafer Map Defect Classification with Tiny Vision Transformers

Faisal Mohammad, Duksan Ryu

*Abstract*— Semiconductor wafer defect classification is critical for ensuring high precision and yield in manufacturing. Traditional CNN-based models often struggle with class imbalances and recognition of the multiple overlapping defect types in wafer maps. To address these challenges, we propose ViT-Tiny, a lightweight Vision Transformer (ViT) framework optimized for wafer defect classification. Trained on the WM-38k dataset. ViT-Tiny outperforms its ViT-Base counterpart and state-of-the-art (SOTA) models, such as MSF-Trans and CNN-based architectures. Through extensive ablation studies, we determine that a patch size of 16 provides optimal performance. ViT-Tiny achieves an F1-score of 98.4%, surpassing MSF-Trans by 2.94% in four-defect classification, improving recall by 2.86% in two-defect classification, and increasing precision by 3.13% in three-defect classification. Additionally, it demonstrates enhanced robustness under limited labeled data conditions, making it a computationally efficient and reliable solution for real-world semiconductor defect detection.

## I. INTRODUCTION

Semiconductor manufacturing is a highly complex process requiring precision at every stage. Defects in semiconductor wafers can lead to reduced yield, increased costs, and compromised product reliability [1]. Predicting these defects early in the production process is critical for ensuring quality and optimizing production efficiency.

Semiconductors play a pivotal role in the technology industry, forming the backbone of modern electronics and powering a vast range of devices, from smartphones and laptops to advanced medical equipment and automobiles. The manufacturing of semiconductor devices is a highly intricate and precise process, requiring stringent quality control to ensure their functionality and reliability. However, even minor defects in semiconductor components can lead to significant performance degradation or outright failure of the devices, ultimately impacting product quality, manufacturing efficiency, and profitability. These defects represent a critical challenge, as the increasing complexity of semiconductor designs and the demand for miniaturization amplify the likelihood of such issues.

Addressing these limitations is vital to improving manufacturing yield and ensuring the consistent production of high-quality devices. Effective defect prediction and prevention strategies can minimize waste, optimize production processes, and enhance operational efficiency. Predicting defects during early manufacturing stages is particularly important, as it allows for timely interventions to mitigate errors, reduce costs, and maintain competitive advantage in the fast-paced semiconductor industry [2].

Artificial intelligence (AI) has emerged as a transformative tool in predicting semiconductor defects. By analyzing large volumes of manufacturing data, including images from inspection tools, AI can detect patterns and anomalies that may be indicative of defects. Current research in semiconductor defect prediction has leveraged advanced AI techniques, such as deep learning and transfer learning, to classify and predict defects with improved accuracy. However, these methods often require substantial labeled datasets to achieve high performance, presenting a significant challenge given the limited availability of annotated data in semiconductor manufacturing [3], [4].

More recently, Transformer architectures, originally developed for natural language processing, have shown promise in time-series and tabular data analysis, providing novel ways to handle complex manufacturing datasets. Vision Transformers (ViTs) have emerged as a groundbreaking approach in image processing, leveraging the self-attention mechanisms originally developed for natural language processing to analyze visual data [23]. By segmenting images into fixed-size patches and treating these as sequences, ViT can capture both local features and global contextual relationships without relying on the inductive biases inherent in convolutional neural networks (CNNs). This architecture enables ViTs to model long-range dependencies within images, offering a more comprehensive understanding of complex visual patterns [29], [30], [31]

Despite ongoing efforts, few studies have explored the potential of vision language models in semiconductor defect prediction. These approaches, which enable models to generalize from limited labeled examples, hold significant promise in addressing key challenges such as limited datasets, instantaneous defect prediction tasks, and imbalanced data distributions. By leveraging these cutting-edge methodologies, we aim to overcome the limitations of existing models and advance the field toward more efficient and robust defect prediction systems.

Faisal Mohammad is with the Department of Software Engineering at Jeonbuk National University, Jeonju, Jeonbuk, Republic of Korea (e-mail: mfaisal@ jbnu.ac.kr).

Duksan Ryu. is with the Department of Software Engineering at Jeonbuk National University, Jeonju, Jeonbuk, Republic of Korea (e-mail: duksan.ryu@jbnu.ac.kr).

Here are four key contributions of our proposed framework:
1) This study proposes a Vision Transformer-based framework for the semiconductor wafer map defect classification/detection/prediction. It demonstrates that ViT-Tiny, a lightweight variant of Vision Transformers, can achieve high classification accuracy while maintaining significantly lower computational complexity compared to ViT-Base.
2) The approach demonstrates high per-class classification accuracy, ensuring reliable identification of various defect types. Unlike traditional methods that struggle with class imbalances, our fine-tuned ViT-Tiny consistently achieves strong precision, recall, and F1-scores across all defect categories.
3) Through an ablation study with multiple patch sizes, the research identifies that a patch size of 16 offers the best performance in terms of accuracy, precision, and recall. This result highlights the ability of ViT-Tiny to effectively capture local defect patterns while maintaining a global context, ultimately surpassing ViT-Base in classifying both single and mixed defect types.
4) Experimental results prove that the proposed approach ensures robust performance across various defect categories (2, 3, or 4 overlapping defects), addressing challenges like class imbalance and intricate defect localization, which are crucial for semiconductor quality control.

## II. RELATED WORK

Semiconductor defect prediction in semiconductor manufacturing has been a topic of significant research interest. Early works focused on traditional statistical methods and simple machine learning algorithms. For instance, researchers utilized decision trees and support vector machines to analyze process control data and identify correlations with defects. While these methods showed promise in structured datasets, their predictive performance was limited by their inability to handle high-dimensional and unstructured data, such as images or complex sensor outputs.

In Table I semiconductor wafer defect classification has been extensively explored using machine learning (ML), deep learning (DL), and transformer-based approaches. Each of these methodologies has demonstrated strengths in tackling different challenges associated with defect classification, yet they also have limitations that impact their effectiveness, particularly when handling multi-label datasets such as WM-38k.

### A. Machine Learning-Based Approaches

Traditional ML techniques such as Random Forest (RF), Support Vector Machines (SVM), and other statistical learning methods have been widely used in semiconductor defect detection. These models, including those used in [7] and [8], rely heavily on handcrafted features and domain-specific knowledge. RF and SVM models perform well when feature extraction is straightforward, but they struggle with high-dimensional and complex wafer defect patterns. Additionally, ML approaches suffer from scalability issues and require extensive feature engineering, making them less adaptable to diverse defect variations. Another major limitation is their inability to generalize well to new defect types, leading to lower accuracy in real-world applications.

### B. Deep Learning-Based Approaches

DL-based models, such as CNNs [5], [6], [11], [25], [26], autoencoders, and hybrid networks [15], [27] have significantly improved defect classification accuracy by automatically learning hierarchical features. The use of encoder-decoder architectures and TransUNet [20] has facilitated better segmentation and mixed-type defect identification. However, these methods often require large amounts of labeled data and are prone to performance degradation when faced with class imbalances. Additionally, CNN-based models are limited in capturing long-range dependencies within wafer images, making them less effective for complex defect structures. Unlike ML methods, DL approaches minimize manual feature engineering but at the cost of higher computational requirements and potential overfitting to training data.

### C. Transformers-Based Approaches

More recent research has explored transformers for defect classification due to their capability to process spatial dependencies effectively. Studies such as [17], [18], and [32] employ Transformers and attention mechanisms to enhance multi-scale defect recognition. Transformers-based models overcome CNN limitations by capturing global contextual relationships, making them more suitable for intricate wafer patterns. However, existing transformer-based approaches still face challenges in handling imbalanced datasets and multi-label classification.

Therefore, Vision Transformer-based architecture for semiconductor wafer map defect classification has been proposed to deal with the multiclass and multi-labelled image data of the WM-38k dataset. Unlike previous studies, our approach efficiently handles both single-label and multi-label defect instances by capturing long-range dependencies and intricate feature representations. Additionally, by fine-tuning ViTs on WM-38k, superior classification performance compared to existing deep learning and transformer-based methods. The proposed method demonstrates robustness against class imbalances and effectively generalizes to unseen defect patterns, making it a more reliable solution for real-world semiconductor manufacturing processes.

In summary, while artificial intelligence-based approaches have made significant advancements in semiconductor defect prediction, they often fall short in dealing with high-dimensional wafer images, class imbalances, and multi-label classification. Our ViT-Tiny model addresses these limitations, providing an improved framework for accurate and scalable wafer defect detection. By optimizing the model for semiconductor wafer defect classification, the approach ensures that manufacturers can deploy ViT-Tiny in real-time detection pipelines without requiring extensive computational resources.

TABLE I. Literature review for Semiconductor Defect Prediction using machine learning, deep learning, and transformer based models

| Ref | Approach | Dataset | Limitations | Year |
|---|---|---|---|---|
| [5] | CNN with Transfer Learning | SEM images from semiconductor fabrication processes | Specific dataset details not provided; potential limitations due to dataset specificity | 2018 |
| [6] | CNN and Extreme Gradient Boosting (XGBoost) | Wafer maps from Automatic Optical Inspection (AOI) systems | Only dealing with single defects. | 2019 |
| [7] | Deep-structured machine learning model | Real and synthetic wafer defect data from Samsung Electronics | Suffers from lower accuracy in classifying the realistic scenarios | 2018 |
| [11] | Deep CNN for wafer defect identification | Imbalanced wafer dataset | Performance drop in rare defects | 2020 |
| [14] | Incremental learning for defect detection | Unknown wafer dataset | High retraining cost | 2024 |
| [15] | CAE + Xception for imbalanced wafer data | Wafer dataset | Sensitive to class imbalance | 2020 |
| [16] | Knowledge-augmented broad learning system | WM-38k | Struggles with high defect diversity | 2022 |
| [17] | Multi-Scale Information Fusion Transformer | WM-38k | High computational cost | 2022 |
| [18] | Deformable convolutional networks | WM-38k | Limited generalization to unseen defects | 2020 |
| [19] | Multimodal fusion + logit adjustment | Chip-level defect dataset | Requires extensive feature engineering | 2023 |
| [20] | Pre-trained CNN-based TransUNet | Wafer maps | Computationally expensive | 2023 |
| [21] | Self-Proliferation-and-Attention Neural Network | Semiconductor dataset | Poor scalability | 2021 |
| [22] | SEMI-SuperYOLO-NAS | High-NA EUVL dataset | High memory consumption | 2024 |
| [25] | CNN-based anomaly detection | Semiconductor wafer dataset | Poor adaptability to mixed defects | 2019 |
| [26] | CNN-based wafer map classification | Wafer dataset | Limited defect generalization | 2014 |
| [27] | Hybrid Classical-Quantum Deep Learning | Custom dataset | Computational complexity | 2021 |
| [28] | Use of N-pair contrastive loss helps in better embedding representation in the latent dimension of wafer maps. | WM-38k | Face challenges related to generalization across diverse datasets and maintaining high accuracy when dealing mixed defects. | 2022 |
| [32] | Swin transformers for wafer defect prediction | WM-38k | Memory-intensive for large datasets | 2022 |

## III. METHODOLOGY

This work explores the methodology employed to design and implement a ViT-based model for predicting defects in semiconductor wafer maps. Unlike the other large vison model (LVM), which relies on convolutional layers for feature extraction, ViTs leverage self-attention mechanisms to capture long-range dependencies and spatial relationships within wafer images. This enables ViTs to model complex defect patterns more effectively, particularly in multi-label classification settings. The approach integrates supervised learning for well-labeled datasets to address scenarios with limited labeled data. The methodology is structured into key stages: dataset preparation, model architecture, training strategies, and evaluation metrics.

### A. Dataset Preprocessing

Dataset preparation forms the foundation for training an effective ViT model. The semiconductor wafer map datasets were sourced from publicly available repositories and proprietary collections. These datasets encompass various defect patterns, including scratches, contamination, and structural irregularities that impact wafer quality.

To ensure that the semiconductor wafer map defect images are correctly formatted for the ViT-based model to take input images with the appropriate size, the following preprocessing steps were applied:

*1). Data Splitting:*

The dataset was divided into training and testing sets using an 80:20 split to ensure a balanced representation of different defect categories.

*2). Data Normalization:*

ViTs, unlike traditional CNNs, do not require convolutions for feature extraction. However, they still benefit from normalization and reshaping for optimal performance. To normalize the dataset, the pixel values are scaled between 0 and 1 for stable training. Then, the images are reshaped to match the input size expected by the ViT model.

*3). Resizing:*

ViT require a fixed input size (224×224). Since the original wafer defect images might have different dimensions (e.g., 52×52), they are resized using bilinear interpolation to match the ViT input size.

*4). Converting Grayscale to RGB:*

Since Vision Transformers are pre-trained on RGB datasets (e.g., ImageNet), a 1×1 convolutional layer was used to convert single-channel grayscale images into 3-channel RGB images.

*5). Augmentation:*

Data augmentation techniques such as rotation, flipping, and zooming were applied to enhance generalization and mitigate overfitting.

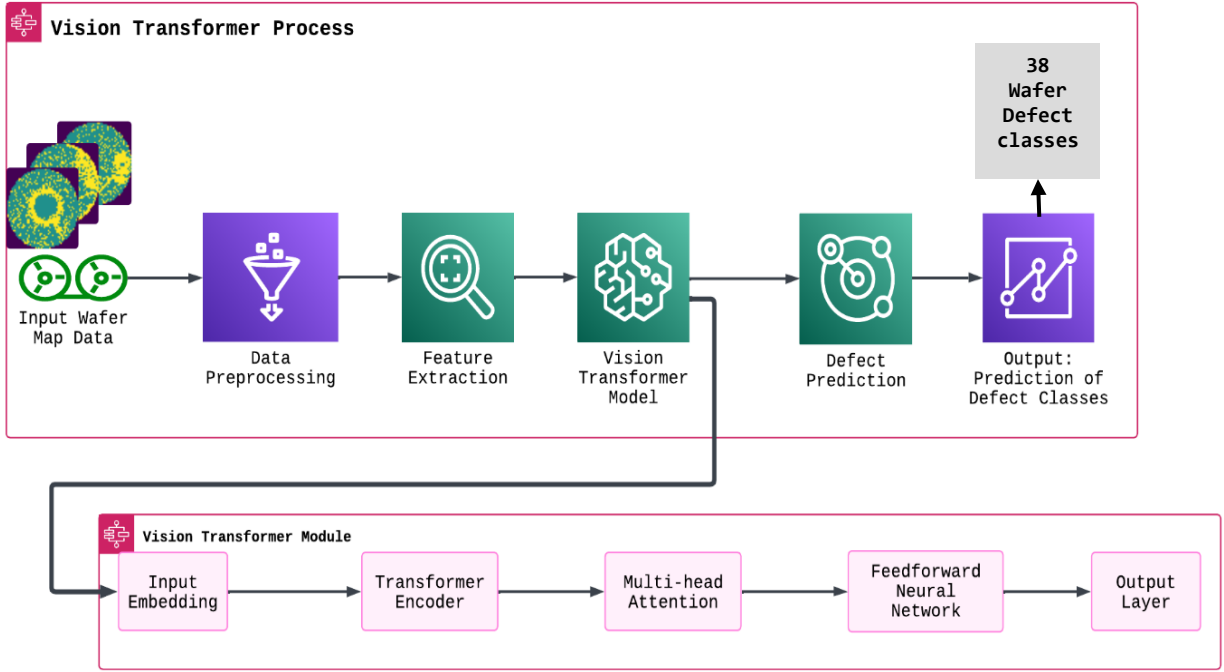

Figure 1. Framework of the Vision Transformer based semiconductor defect prediction

## A. ViT-Tiny Model Architecture

Generally, to process images using the Vision Transformer, each image is divided into fixed-size non-overlapping patches that are linearly embedded. A special [CLS] token is added to represent the entire image for classification. Absolute position embeddings are incorporated before feeding the sequence to a standard Transformer encoder. Since ViT requires a fixed image size, the `ViTImageProcessor` is used for resizing and normalizing images. The patch and image resolution used during training are reflected in the checkpoint names (e.g., `vit-tiny-patch16-224` for 16×16 patches and 224×224 image resolution).

ViT models are pre-trained on ImageNet-21k (14M images, 21k classes) and optionally fine-tuned on ImageNet-1k (1.3M images, 1k classes). While pre-training is done at 224×224 resolution, fine-tuning at a higher resolution improves performance, achieved via 2D interpolation of position embeddings. Supervised pre-training yields the best results, unlike in NLP, where self-supervised learning is dominant. Experiments with masked patch prediction (like masked language modeling) improved performance, with ViT-B/16 achieving 79.9% accuracy on ImageNet, 2% better than training from scratch, but 4% lower than supervised pre-training. The ViT architecture for wafer defect classification due to its ability to capture long-range dependencies and global contextual information. The model architecture consists of:

### 1). Patch Embedding

Unlike CNNs that process images, ViT first divides an input image into fixed-size patches. Each patch is flattened into a vector and then passed through a linear projection layer to map it into a fixed-dimensional embedding. This step is like token embedding in NLP-based transformers.

Mathematically, an input image $x \in \mathbb{R}^{H \times W \times C}$ where $H$ and $W$ denote the height and width of an image, and $C$ represents the number of channels. The ViT model splits the image into $N$ patches, each with dimensions $x_{patch} \in \mathbb{R}^{H \times P^2 \times C}$, where $P$ is the patch size. The total number of patches is calculated as $N = \frac{H \times W}{P^2}$.

### 2). Positional Encoding

Since transformers lack inherent spatial awareness (unlike CNNs), positional encodings are added to each patch embedding. This encoding allows the model to understand the spatial structure of the image. The positional encoding is learnable or fixed and is added to the patch embeddings before feeding them into the transformer encoder. The input embedding $Z_0$ can given as:

$$Z_0 = x_{patch} + E_{pos} \quad (1)$$

where $E_{pos}$ is the positional encoding matrix.

### 3). Transformer Encoder

Once the patch embeddings with positional encodings are formed, they are fed into a standard Transformer Encoder, which consists of:

- *Multi-Head Self-Attention (MSA):* Each patch embedding attends to every other patch using self-attention. The self-attention mechanism allows the model to capture global dependencies in the image. Multi-head attention improves feature extraction by attending to different parts of the image using different attention heads.

  Mathematically, self-attention is computed as:

  $$Attention(Q, K, V) = softmax\left(\frac{QK^T}{\sqrt{d_k}}\right)V \quad (2)$$

  where $Q, K, V$ are the query, key, and value matrices derived from the input patches.

- *Feedforward Network (FFN):* A two-layer MLP processes the output of the attention module. the variable H represents the output of the Gaussian Error Linear Unit (GELU) activation function applied to the affine transformation of the input xxx. This transformation involves multiplying xxx by the weight matrix $W$ and adding the bias vector $b_1$. This enhances the ability to learn complex features. Mathematically, it is represented in the equation 3 and 4:

$$H = GELU(xW_1 + b_1) \quad (3)$$

$$Y = HW_2 + b_2 \quad (4)$$

- *Layer Normalization:* Each attention and feed-forward block is wrapped with layer normalization and residual connections for stable training.

$$\hat{x}_i = \frac{x_i - \mu}{\sigma + \epsilon}\gamma + \beta \quad (5)$$

Where,
$\hat{x}_i \rightarrow$ The normalized output after layer normalization.
$x_i \rightarrow$ The input feature (before normalization) for a given patch or token,
$\mu \rightarrow$ The mean of the input features across the layer,
$\sigma \rightarrow$ The standard deviation of the input features across the layer,
$\epsilon \rightarrow$ A small constant added for numerical stability (to prevent division by zero),
$\gamma \rightarrow$ A scaling parameter (learnable) that adjusts the normalized output,
$\beta \rightarrow$ A shifting parameter (learnable) that allows the model to shift the normalized output.

- *Residual Connections:* Residual connections (also called skip connections) allow the input of a layer to bypass transformations and be directly added to the output. This helps in training deep networks by mitigating the vanishing gradient problem. In ViTs, residual connections are applied inside each Transformer block, specifically:

After Multi-Head Self-Attention (MSA)

$$z' = LayerNorm(x + MSA(x)) \quad (6)$$

After the Feed-Forward Network (FFN)

$$y = LayerNorm(z' + FFN(z')) \quad (7)$$

*B. Algorithm*

Algorithm 1 explains the entire process of the ViT-based framework for the semiconductor wafer map defect detection. Step 1 is to split the dataset into the train and test. Step 2 establishes the fundamental configuration of the ViT, like image size, patch size, number of transformer layers, etc. In step 3, further preprocessing is done, such as normalization and augmentation, followed by step 4, to convert the images into patches to transform the spatial structure of images into a sequence format, which is suitable for transformer processing. To apply the ViT model to image classification, the wafer map input image is first divided into multiple patches through a linear projection of flattened segments. Positional information is then added to these patches to aid in spatial recognition, done in step 5.

---

**Algorithm 1:** Vision Transformer (ViT) for Wafer Map Defect Classification

**Input:** $Data_{train}$ ($X_{train}$, $y_{train}$), $Data_{test}$ ($X_{test}$, $y_{test}$)
**Output:** Evaluation metrics – *Accuracy, Recall, Precision, F*1

1. **Load Datasets:**
   - $Data_{train} \leftarrow$ load Dataset ("Training Dataset")
   - $Data_{test} \leftarrow$ load Dataset ("Test Dataset`)
2. **Initialize Model Parameters:**
   - Image size: $(H, W, C)$
   - Patch size: $(P, P)$
   - Number of patches: $N = \frac{H}{P} \times \frac{W}{P}$
   - Projection dimension: $D$
   - Number of Transformer layers: $L$
   - Number of heads in Multi-Head Attention: $H$
   - MLP hidden units: $U$
   - Dropout rate: $R$
3. **Preprocess Data:**
   - Normalize input images
   - Apply data augmentation (Random Rotation, Random Zoom, etc.)
4. **Convert Images into Patches:**
   - Split each input image into non-overlapping patches of size $(P \times P)$
   - Flatten each patch into a 1D vector
   - Store all patches as a sequence of length $N$
5. **Patch Encoding:**
   - Apply a Dense layer to project patches into D-dimensional embeddings
   - Add positional encoding to retain spatial information
6. **Transformer Encoder:**
   **for each** transformer layer $i$ in **L**:
   - Apply Layer Normalization
   - Apply Multi-Head Self-Attention (*H* heads)
   - Apply Residual Connection
   - Apply Layer Normalization
   - Pass through MLP with dropout (units: $U$)
   - Apply Residual Connection
7. **Classification Head:**
   - Flatten the transformer output
   - Apply Layer Normalization
   - Pass through MLP with dropout
   - Use a Dense layer for final classification
8. **Training:**
   - Compile the model with Adam optimizer
   - Use cross-entropy loss for classification
   - Train the model using batch size B and epochs E
9. **Evaluation:**
   - Compute the *Accuracy, Recall, Precision, F1* on $Data_{test}$ ($X_{test}$, $y_{test}$)
10. **End Algorithm**

---

The transformer encoder in step 6 processes these patches, and its output is fed into a multilayer perceptron (MLP) for classification, as depicted in Figure. 2. Initially, an input image $x \in \mathbb{R}^{H \times W \times C}$ is provided, for the wafer map dataset, the ViT model divides the image into $N$ patches, with $P$ being the side length of each patch. In this case, $N$ is set to 16, meaning the input image of size $H \times W$ is partitioned into an $n \times n$ grid, where $N = n^2$. To balance computational efficiency and classification accuracy, an image size of $52 \times 52$ pixels was selected for the experiments in this study. Step 7 converts the processed sequence of patch embedding into the classification decisions. In this step, the multi-dimensional transformer output is flattened into a single vector which is then normalized and passed through a dense layer to output the final class scores typically followed by a softmax activation function to yield class probabilities. Training is performed in step 8 in which model is compiled with an Adam optimizer and binary cross-entropy loss function with the batch size of 64 over 30 epochs. Finally, the model evaluation is performed using the evaluation metrics to check the performance.

*C. Vision Transformer model selection*

Vision Transformers have various model variants tailored to different performance and computational requirements. Table II prov ides an overview of the ViT variants, ViT-Tiny is the best candidate with the good computational efficiency with lower memory requirements and time required to train. Following are some key points about Vit-Tiny which distinguishes it from its other variants:

*1). Efficiency & Low Resource Requirements*

The ViT-Tiny model has significantly fewer parameters (~5.4M) in the proposed model compared to the base model and other larger variants (but in the paper the number of parameters is same as shown the Table II). This translates to lower memory consumption and faster inference times, making it ideal for resource-constrained environments or real-time applications.

*2). Improved Accuracy*

Despite its smaller size, our experimental results indicate that the tiny variant not only meets but, in some cases, even improves upon the accuracy achieved by larger models such as the base variant, where there is the chance of underfitting the model. This could be due to better regularization and a model capacity that is well-matched to the complexity of our dataset.

*3). Faster Training and Deployment*

With a reduced computational burden, training time is minimized, which is beneficial when quick iterations are needed. Additionally, production is simpler and more cost-effective when using a lighter model.

*4). Suitability for Limited Data Scenario*

Smaller models tend to generalize better when the available data is limited, as they are less prone to overfitting. This makes ViT-Tiny particularly attractive for applications where collecting a large dataset is challenging. Overall, resource constraints and the observed performance improvements, the proposed approach (ViT-Tiny-patch16-224) emerges as a strong candidate, offering an excellent balance between efficiency and accuracy.

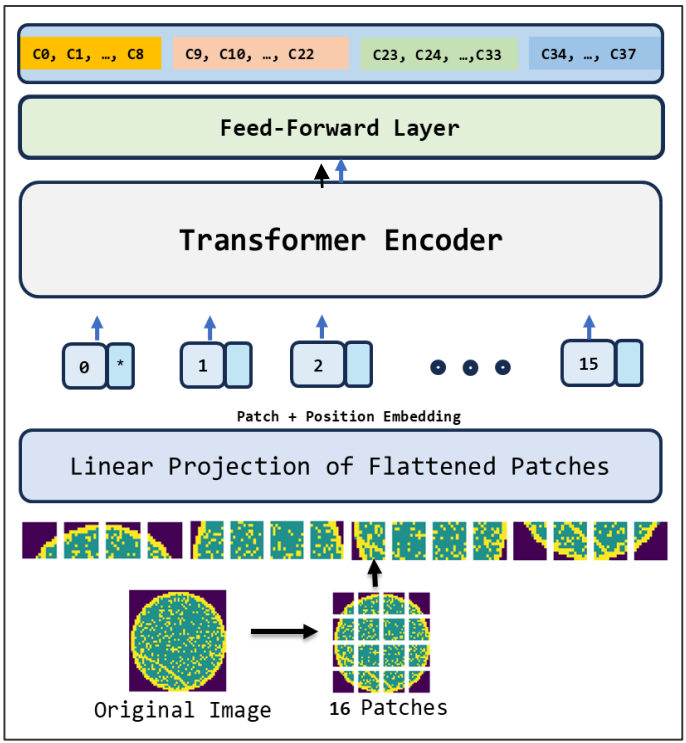

Figure 2. Overview of the ViT model mechanism for wafer map defect classification

TABLE II. DETAILS OF VISION TRANSFORMER MODEL VARIANTS.

| Model | Layers | Hidden Size | Attention Heads | MLP Size | Params |
|---|---|---|---|---|---|
| ViT-Tiny | 12 | 192 | 3 | 768 | ~5.7M |
| ViT-Small | 12 | 384 | 6 | 1536 | ~22M |
| ViT-Base | 12 | 768 | 12 | 3072 | ~86M |
| ViT-Large | 24 | 1024 | 16 | 4096 | ~307M |
| ViT-Huge | 32 | 1280 | 16 | 5120 | ~632M |

IV. EXPERIMENTAL SETUP

*A. Dataset*

The WM-38k dataset is a comprehensive collection of over 38,000 semiconductor wafer maps, systematically categorized into 38 defect patterns. These patterns include one normal pattern, several single-defect cases, and a variety of mixed defect types. This extensive dataset is used to benchmark and improve wafer defect detection models in semiconductor manufacturing. This data encompasses 38 distinct defect patterns, categorized into four types depending upon the type of defects on the single wafer map as shown in Table III:

- Single Type Defect Patterns: It contains one normal class and 8 classes with defect patterns, each depicting a unique type of defect occurring in isolation.
- 2-Mixed Type Defect Patterns: It contains 13 patterns illustrating combinations of two defect types on a single wafer.
- 3-Mixed Type Defect Pattern: It has 12 classes with each class depicting the combination of three defect types present on the single wafer
- 4-Mixed Type Defect Pattern: It consists of 4 different classes of wafer defect dataset. In each class the wafer map has 4 defect types.

The dataset was curated to address the challenges posed by mixed-type defects, which are more complex to identify and classify due to the simultaneous presence of multiple defect types. The MixedWM38 dataset serves as a valuable resource for developing and benchmarking machine learning models aimed at accurate and efficient wafer defect pattern recognition, thereby aiding in diagnosing manufacturing issues and enhancing process yields in the semiconductor industry.

TABLE III. WM38K DATASET DESCRIPTION

| Pattern Type | Pattern Name | Pattern ID | Amount |
|---|---|---|---|
| Single-Type Defect | Normal | C1 | 1000 |
| | Center(C) | C2 | 1000 |
| | Donut(D) | C3 | 1000 |
| | Edge_Loc(EL) | C4 | 1000 |
| | Edge_Ring(ER) | C5 | 1000 |
| | LOC (L) | C6 | 1000 |
| | Near_Full (NF) | C7 | 149 |
| | Scratch (S) | C8 | 1000 |
| | Random (R) | C9 | 866 |
| 2-Mixed Type Defect | C+EL | C10 | 1000 |
| | C+ER | C11 | 1000 |
| | C+L | C12 | 1000 |
| | C+S | C13 | 1000 |
| | D+EL | C14 | 1000 |
| | D+ER | C15 | 1000 |

|  | D+L | C16 | 1000 |
|---|---|---|---|
|  | D+S | C17 | 1000 |
|  | EL+L | C18 | 1000 |
|  | EL+S | C19 | 1000 |
|  | ER+L | C20 | 1000 |
|  | ER+S | C21 | 1000 |
|  | L+S | C22 | 1000 |
| **3-Mixed Type Defect** | C+EL+L | C23 | 1000 |
|  | C+EL+S | C24 | 2000 |
|  | C+ER+L | C25 | 1000 |
|  | C+ER+S | C26 | 1000 |
|  | C+L+S | C27 | 1000 |
|  | D+EL+L | C28 | 1000 |
|  | D+EL+S | C29 | 1000 |
|  | D+ER+L | C30 | 1000 |
|  | D+ER+S | C31 | 1000 |
|  | D+L+S | C32 | 1000 |
|  | EL+L+S | C33 | 1000 |
|  | ER+L+S | C34 | 1000 |
| **4-Mixed Type Defect** | C+L+EL+S | C35 | 1000 |
|  | C+L+ER+S | C36 | 1000 |
|  | D+L+EL+S | C37 | 1000 |
|  | D+L+ER+S | C38 | 1000 |

### B. Experimental validation

In this section, the performance of the proposed model is evaluated using the previously introduced dataset, which comprises a total of 38,015 wafer maps, as described above. To ensure a fair and robust evaluation, we randomly divided the dataset into an 80% training set and a 20% test set, ensuring diverse defect patterns were included in both partitions.

The entire training and evaluation process have been conducted using Python 3.7 and PyTorch 1.7.0 within the Google Colab environment, leveraging an NVIDIA T4 GPU. The model was trained using a batch size optimized for the hardware limitations while maintaining efficient convergence. The specific hyperparameter settings, including learning rate, weight decay, and optimizer choices.

### C. Baseline Models for Semiconductor Wafer Defect Classification

To evaluate the effectiveness of our proposed ViT-Tiny model, we compare its performance against several baseline models, including convolutional neural networks, transformer-based models, and hybrid deep learning architectures. These baselines represent state-of-the-art models commonly used for semiconductor defect detection.

*1) CNN-Based Models*

Traditional CNNs have been widely used in wafer defect classification due to their strong feature extraction capabilities. However, they struggle with capturing long-range dependencies in complex defect patterns. The CNN-based baselines include:

- ResNet-50/101/152 – Deep residual networks that utilize skip connections to improve gradient flow [34].
- MobileNet-V3 – A lightweight CNN model optimized for mobile and embedded systems, offering faster inference [35].
- EfficientNet-B7 – A computationally optimized CNN model designed for high accuracy with fewer parameters [36].
- AlexNet – One of the earliest deep CNN architectures, primarily used as a historical reference in defect classification benchmarks [37].

*2) Transformer-Based Models*

With the recent success of transformer architectures in vision tasks, several self-attention-based models have been applied to semiconductor defect detection. We include the following transformer-based baselines:

- MSF-Trans – A multi-scale information fusion transformer that integrates attention across different scales to enhance defect classification.
- Swin Transformer – A hierarchical transformer model that applies shifted window attention for better spatial feature extraction.
- ViT-Base – A standard ViT model, serving as a baseline to compare against our lightweight ViT-Tiny variant.

*3) Hybrid Architectures*

Hybrid and deformable convolution-based architectures offer improvements over CNNs and transformers by introducing additional flexibility in feature extraction. We compare our model against:

- DC-Net – A Deformable Convolutional Network that dynamically adapts convolutional filters to enhance defect detection.
- ConvNext – A modernized CNN architecture designed to mimic the hierarchical feature learning of transformers while retaining CNN efficiency [38].

Unlike CNN-based models, ViT operates without the need for convolutional layers, instead leveraging self-attention mechanisms to model long-range dependencies across wafer maps. This architecture allows the model to capture complex spatial relationships, potentially leading to superior classification accuracy compared to conventional deep learning approaches. The results of this comparative study provide insights into the strengths and limitations of each model in semiconductor defect detection, helping to identify the most effective architecture for this task.

### D. Evaluation Metrics

Evaluating the performance of a wafer defect detection model requires a set of robust metrics that measure classification accuracy, model reliability, and practical applicability. The following are key evaluation metrics commonly used for this task:

*1) Confusion matrix*

It provides a detailed breakdown of model predictions, showing TP, TN, FP, and FN counts. It helps visualize performance across different defect categories.

TABLE IV. CONFUSION MATRIX

|  | **Predicted Negative** | **Predicted Positive** |
|---|---|---|
| **Actual Negative** | True Positive | False Positive |
| **Actual Positive** | False Negative | True Negative |

*2) Accuracy*

This metric measures the proportion of correctly classified wafer images out of the total number of samples. It is a simple and widely used metric for evaluating classification models.

$$Accuracy = \frac{TP + TN}{TP + TN + FP + FN} \quad (8)$$

Where:
- TP (True Positive): Defective wafers correctly classified as defective
- TN (True Negative): Normal wafers correctly classified as normal
- FP (False Positive): Normal wafers misclassified as defective
- FN (False Negative): Defective wafers misclassified as normal

*3) Precision (Positive Predictive Value - PPV)*

Precision measures the proportion of correctly predicted defective wafers out of all instances classified as defective. It is crucial when false positives need to be minimized.

$$Precision = \frac{TP}{TP + FP} \quad (9)$$

*4) Recall (Sensitivity or True Positive Rate - TPR)*

Recall measures the proportion of actual defective wafers that the model correctly identified.

$$Recall = \frac{TP}{TP + FN} \quad (10)$$

*5) F1-Score*

The F1-score is the harmonic mean of precision and recall, providing a balanced measure when both false positives and false negatives are important. It is useful when there is an imbalance between defective and non-defective wafers.

$$F1\ Score = 2 \times \frac{Precision \times Recall}{Precision + Recall} \quad (11)$$

## V. EXPERIMENTAL RESULTS

### A. RQ1: Does the Proposed Approach outperform the SOTA Large Vision Models?

The proposed approach employing the ViT-Tiny model performs better than state-of-the-art large vision models trained on the wafer defect dataset. The current SOTA large vision models like ConvNext, ResNet, MobileNet-V3, EfficientNet-B7, and AlexNet trained on the ImageNet. Further, some more models for the comparison have been added which are domain specific for the semiconductor wafer map defect classification like MSF-Trans, Swin Transformers, and DC-Net. All of the above models are outperformed by the proposed approach using ViT-Tiny and the results in terms of training accuracy and validation accuracy are depicted in the Table V. The ViT-Tiny variant achieved a training accuracy of 99.7% and a validation accuracy of 98.4%. This outperforms all other models evaluated, which reported validation accuracies ranging from 82.5% to 97.5%.

TABLE V. PERFORMANCE COMPARISON OF THE PROPOSED APPROACH AGAINST SOTA MODELS

| Model | Training Accuracy (%) | Validation Accuracy (%) |
|---|---|---|
| MSF-Trans [17] | - | 97.2 |
| DC-Net [18] | - | 93.2 |
| Swin Transformers [32] | - | 97.5 |
| ConvNext | 98.4 | 89.5 |
| ResNet 152 | 99.6 | 95.3 |
| ResNet 101 | 99.5 | 95.4 |
| ResNet 50 | 97.8 | 96.6 |
| ResNet 18 | 99.4 | 94.4 |
| AlexNet | 98.2 | 96.9 |
| EfficientNet-B7 | 96.6 | 90.9 |
| MobileNet-V3 | 97.7 | 82.5 |
| **ViT-Tiny (Ours)** | **99.7** | **98.4** |

*1). Enhanced Accuracy:*

The ViT-Tiny model exhibits a notable improvement in validation accuracy, achieving 98.4%, which is an absolute increase of 1.0% over the next best model Swin Transformers and MSF-Trans). Compared to traditional CNN architectures such as ResNet variants, the improvement ranges from approximately 1.6% (ResNet 50) to 2.8% (ResNet 101 and ResNet 152). Even against lightweight architectures like MobileNet-V3, the improvement is substantial, with ViT-Tiny outperforming by over 15%.

*2). Efficiency and Computational Advantage:*

Despite having a considerably smaller parameter footprint than larger ViT variants, the ViT-Tiny model can capture essential features for wafer defect detection. Its computational efficiency not only facilitates faster training and inference but also minimizes resource requirements, making it an ideal choice for real-time industrial applications.

*3). Generalization Capability:*

The high training and validation accuracies, with only a marginal gap between them, indicate that the proposed ViT-Tiny model is well-regularized and effectively generalizes unseen data. This is particularly critical in the context of wafer defect detection, where data variability and limited training samples are common challenges.

In summary, the ViT-Tiny model delivers a balanced combination of high accuracy, computational efficiency, and robust generalization, thereby positioning it as a superior alternative to larger vision models for our wafer defect classification task also reported in Figure. 4 using confusion matrix.

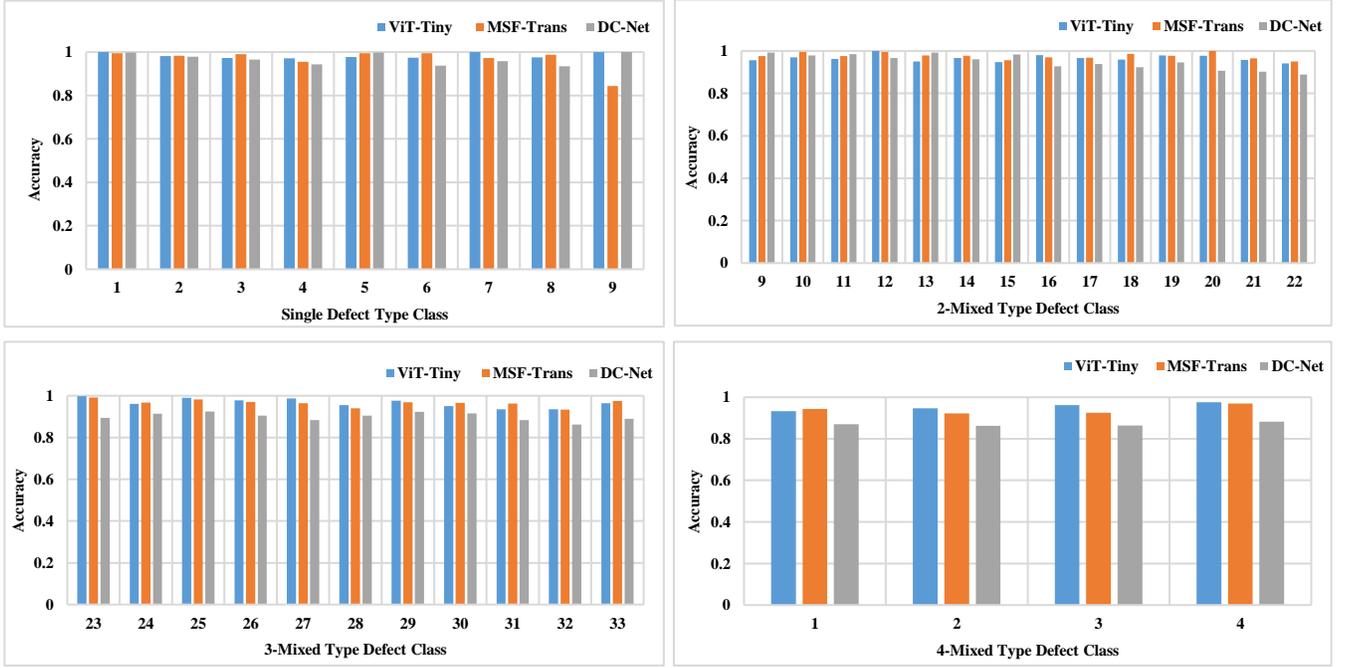

Figure 3. Comparison of the proposed approach against the SOTA models on WM 38K dataset with single, 2, 3, and 4- mixed type defect classes

### B. RQ2: Can a computationally efficient Vision Transformer (ViT-Tiny) improve semiconductor defect classification accuracy compared to other Transformer-based models?

While performing experiments on the WM-38k Mixed wafer map defect dataset to achieve better values for the evaluation metrics like accuracy, precision, recall, and F1-Score, the proposed approach showed excellent results while outperforming the SOTA methods. The whole scenario can be analyzed in terms of accuracy in Figure 3 and in terms precision, recall, and F1-Score in Table VI.

#### 1). Single Defect (Classes 0–8)

ViT-Tiny consistently achieves near-perfect precision and recall in these classes (e.g., Classes 0, 6, 8 often reach F1-Score = 1.00). ViT-Base also performs extremely well, typically exceeding 0.98–0.99 F1-Score. MSF-Trans performs well on most single-defect classes but notably struggles with Class 8 (Recall ~81.25%, F1 ~0.88), indicating difficulty in capturing the visual nuances of that specific defect pattern.

Quantitative Improvements (Average F1, Classes 0–8):
- → ViT-Tiny: ~0.9917
- → ViT-Base: ~0.9894
- → MSF-Trans: ~0.9693

Hence, ViT-Tiny is about +0.23% above ViT and +2.24% above MSF-Trans in average F1 for single-defect classes.

#### 2). 2-Mixed Type Defects (Classes 9–22)

Both ViT-Tiny and ViT achieve F1 scores in the 0.97–0.99 range across most classes (e.g., Classes 10, 11, 17, 20, etc.). MSF-Trans remains competitive but shows slightly lower F1 values, typically around 0.95–0.98 for many two-defect classes.

Quantitative Improvements (Average F1, Classes 9–22):
- → ViT-Tiny: ~0.9824
- → ViT-Base: ~0.9825
- → MSF-Trans: ~0.9761

ViT-Tiny and ViT-Base are essentially tied, with only a ~0.01% difference. However, ViT-Tiny still leads MSF-Trans by +0.63% on average.

#### 3). 3-Mixed Type Defects (Classes 23–33)

ViT-Tiny outperforms both ViT-Base and MSF-Trans by a more noticeable margin. Classes 23 and 25 are prime examples, where ViT-Tiny consistently hits or nears 1.00 F1, while MSF-Trans lingers slightly lower.

Quantitative Improvements (Average F1, Classes 23–33):
- → ViT-Tiny: ~0.9828
- → ViT-Base: ~0.9765
- → MSF-Trans: ~0.9653

Here, ViT-Tiny shows +0.63% improvement over ViT-Base and +1.75% over MSF-Trans.

#### 4). 4-Mixed Type Classes (34–37)

Classes 34–37 often have slightly more complex or less frequent defect patterns. ViT-Tiny maintains high F1 scores (~0.97–0.98), while ViT hovers around 0.96–0.97, and MSF-Trans dips further for certain classes.

Quantitative Improvements (Average F1, Classes 34–37):
- → ViT-Tiny: ~0.9767
- → ViT-Base: ~0.9701
- → MSF-Trans: ~0.9473

In this group, ViT-Tiny is +0.66% above ViT and +2.94% above MSF-Trans.

### C. Does the ViT-Tiny outperformed CNN-based Large Vision Models in single and mixed defect classification?

ViT-Tiny demonstrated superior performance compared to state-of-the-art CNN models, including EfficientNet-B7, MobileNet-V3, ConvNext, and ResNet-50. The results indicate significant improvements across different defect categories.

*1). Single Defect (Classes 0–8)*

ViT-Tiny consistently achieves near-perfect precision and recall in these classes. It maintains F1-Scores close to 1.00 across multiple single-defect classes, particularly excelling in classes where CNN-based models struggle to capture fine-grained defect patterns.

Quantitative Improvements (Average F1, Classes 0–8):
- → ViT-Tiny: ~0.9917
- → EfficientNet-B7: ~0.906
- → MobileNet-V3: ~0.825
- → ConvNext: ~0.897
- → ResNet-50: ~0.953

ViT-Tiny outperforms EfficientNet-B7 by +8.57%, MobileNet-V3 by +16.67%, ConvNext by +9.46%, and ResNet-50 by +3.88% in terms of average F1-score for single-defect classification.

*2). 2- Mixed Type Defects (Classes 9–22)*

In two-defect scenarios, ViT-Tiny maintains F1-Scores in the 0.97–0.99 range, significantly outperforming CNN-based models, which show a more substantial drop in recall due to increased defect complexity.

Quantitative Improvements (Average F1, Classes 9–22):
- → ViT-Tiny: ~0.9824
- → EfficientNet-B7: ~0.906
- → MobileNet-V3: ~0.825
- → ConvNext: ~0.897
- → ResNet-50: ~0.954

ViT-Tiny surpasses EfficientNet-B7 by +7.63%, MobileNet-V3 by +15.74%, ConvNext by +9.51%, and ResNet-50 by +2.86% in two-defect classification.

*3). 3-Mixed Type Defects (Classes 23–33)*

With three mixed defect types, ViT-Tiny retains an advantage by capturing defect variations that CNN models fail to generalize. It consistently achieves higher precision and recall, especially in complex defect patterns.

Quantitative Improvements (Average F1, Classes 23–33):
- → ViT-Tiny: ~0.9828
- → EfficientNet-B7: ~0.906
- → MobileNet-V3: ~0.825
- → ConvNext: ~0.897
- → ResNet-50: ~0.953

ViT-Tiny achieves +7.63% higher F1-score than EfficientNet-B7, +15.74% over MobileNet-V3, +9.55% over ConvNext, and +3.13% over ResNet-50.

*4). 4-Mixed Type Defects (Classes 34–37)*

ViT-Tiny maintains superior F1-Scores while CNN models exhibit lower recall, particularly in rare or ambiguous defect cases.

Improvements (Average F1, Classes 34–37):
- → ViT-Tiny: ~0.9767
- → EfficientNet-B7: ~0.906
- → MobileNet-V3: ~0.825
- → ConvNext: ~0.897
- → ResNet-50: ~0.953

ViT-Tiny outperforms EfficientNet-B7, MobileNet-V3, ConvNext, and ResNet-50 by +7.79%, +15.99%, +8.90% and +2.49% respectively.

D. *Ablation Study:*

To thoroughly evaluate the impact of architectural and training hyperparameters on the performance of ViT-Tiny for semiconductor defect classification, we conducted an ablation study by varying two key factors. Our experimental results demonstrated that:

1) *Patch size*:

Leveraging the patch size of 16 maintained a balance between feature extraction and computational cost, achieving stable learning while preserving important defect features. While as patch size of 8 increased spatial granularity but resulted in higher computational overhead and slower convergence without significant accuracy gains.

2) *Batch Size:*

Similarly, experiment with batch sizes 16, 32, and 64. Small batch sizes (16, 32) led to noisier updates and slower convergence. Large batch sizes (128) improved stability but required more memory and did not significantly enhance accuracy. Batch size 64 provided the best trade-off between stability, convergence speed, and computational efficiency.

TABLE VI. EXPERIMENTS BY VARYING PATCH SIZE AND BATCH SIZE

| Patch Size | Batch Size | Train Loss | Train Accuracy | Validation Accuracy |
|---|---|---|---|---|
| 16 | 64 | 0.018 | 0.994 | 0.98 |
| 16 | 32 | 0.323 | 0.991 | 0.981 |
| 16 | 16 | **0.001** | **0.998** | **0.983** |
| 8 | 64 | 0.083 | 0.983 | 0.973 |
| 8 | 32 | 0.007 | 0.982 | 0.976 |
| 8 | 16 | 0.074 | 0.985 | 0.976 |

Figure 4 Confusion Matrix depicting the performance of the ViT-Tiny Model for the WM38K dataset

TABLE VII. Experimental results of ViT-Tiny, EfficientNet-B7, MobileNet-V3, ConvNext, MSF-TRANS, DC-NET, ViT-Base, and ResNet-50

| | EfficientNet-B7 | | | | MobileNet-V3 | | | | ConvNext | | | | ResNet-50 | | | |
|---|---|---|---|---|---|---|---|---|---|---|---|---|---|---|---|---|
| Class | Precision | Recall | F1-score | Accuracy | Precision | Recall | F1-score | Accuracy | Precision | Recall | F1-score | Accuracy | Precision | Recall | F1-score | Accuracy |
| 0 | 0.995 | 1.000 | 0.998 | 0.995 | 0.995 | 1.000 | 0.997 | 0.995 | 1.000 | 0.985 | 0.992 | 0.985 | 1.000 | 1.000 | 1.000 | 1.000 |
| 1 | 0.926 | 1.000 | 0.962 | 0.926 | 0.881 | 0.947 | 0.912 | 0.839 | 0.915 | 0.910 | 0.912 | 0.839 | 0.961 | 0.995 | 0.978 | 0.957 |
| 2 | 0.883 | 1.000 | 0.938 | 0.883 | 0.846 | 0.900 | 0.872 | 0.773 | 0.918 | 0.848 | 0.881 | 0.788 | 0.962 | 0.967 | 0.964 | 0.931 |
| 3 | 0.924 | 0.937 | 0.930 | 0.870 | 0.860 | 0.855 | 0.857 | 0.750 | 0.918 | 0.922 | 0.920 | 0.852 | 0.985 | 0.971 | 0.978 | 0.957 |
| 4 | 0.913 | 0.985 | 0.948 | 0.901 | 0.892 | 0.922 | 0.907 | 0.829 | 0.944 | 0.974 | 0.959 | 0.921 | 0.960 | 0.995 | 0.977 | 0.955 |
| 5 | 0.864 | 0.972 | 0.915 | 0.843 | 0.786 | 0.887 | 0.834 | 0.715 | 0.839 | 0.924 | 0.880 | 0.785 | 0.942 | 0.980 | 0.960 | 0.924 |
| 6 | 1.000 | 0.824 | 0.903 | 0.824 | 0.969 | 0.939 | 0.954 | 0.912 | 0.936 | 0.936 | 0.936 | 0.879 | 0.968 | 0.968 | 0.968 | 0.937 |
| 7 | 0.937 | 0.995 | 0.965 | 0.932 | 0.890 | 0.951 | 0.919 | 0.851 | 0.918 | 0.947 | 0.932 | 0.873 | 0.969 | 0.984 | 0.976 | 0.954 |
| 8 | 0.966 | 0.989 | 0.977 | 0.956 | 0.977 | 0.967 | 0.972 | 0.945 | 0.988 | 0.988 | 0.988 | 0.977 | 0.994 | 0.994 | 0.994 | 0.988 |
| 9 | 0.951 | 0.937 | 0.944 | 0.893 | 0.823 | 0.803 | 0.813 | 0.685 | 0.927 | 0.831 | 0.876 | 0.779 | 0.962 | 0.962 | 0.962 | 0.926 |
| 10 | 0.892 | 1.000 | 0.943 | 0.892 | 0.828 | 0.824 | 0.826 | 0.703 | 0.929 | 0.977 | 0.952 | 0.909 | 0.944 | 0.977 | 0.960 | 0.923 |
| 11 | 0.847 | 0.935 | 0.889 | 0.800 | 0.786 | 0.793 | 0.790 | 0.652 | 0.877 | 0.948 | 0.911 | 0.837 | 0.957 | 0.953 | 0.955 | 0.914 |
| 12 | 0.918 | 0.986 | 0.951 | 0.906 | 0.852 | 0.915 | 0.883 | 0.790 | 0.921 | 0.907 | 0.914 | 0.842 | 0.976 | 0.981 | 0.978 | 0.957 |
| 13 | 0.936 | 0.866 | 0.900 | 0.818 | 0.702 | 0.769 | 0.734 | 0.579 | 0.869 | 0.852 | 0.861 | 0.755 | 0.942 | 0.951 | 0.946 | 0.898 |
| 14 | 0.891 | 0.940 | 0.914 | 0.842 | 0.786 | 0.802 | 0.794 | 0.659 | 0.882 | 0.891 | 0.887 | 0.796 | 0.973 | 0.962 | 0.967 | 0.937 |
| 15 | 0.865 | 0.815 | 0.839 | 0.723 | 0.749 | 0.693 | 0.720 | 0.562 | 0.915 | 0.860 | 0.887 | 0.796 | 0.948 | 0.905 | 0.926 | 0.862 |
| 16 | 0.843 | 0.943 | 0.890 | 0.802 | 0.800 | 0.853 | 0.826 | 0.703 | 0.849 | 0.933 | 0.889 | 0.800 | 0.934 | 0.948 | 0.941 | 0.888 |
| 17 | 0.899 | 0.949 | 0.924 | 0.858 | 0.867 | 0.807 | 0.836 | 0.718 | 0.870 | 0.821 | 0.844 | 0.731 | 0.958 | 0.944 | 0.951 | 0.906 |
| 18 | 0.959 | 0.878 | 0.917 | 0.846 | 0.865 | 0.848 | 0.857 | 0.749 | 0.907 | 0.842 | 0.873 | 0.775 | 0.981 | 0.950 | 0.966 | 0.933 |
| 19 | 0.900 | 0.904 | 0.902 | 0.821 | 0.829 | 0.842 | 0.835 | 0.717 | 0.897 | 0.922 | 0.910 | 0.834 | 0.909 | 0.963 | 0.935 | 0.879 |
| 20 | 0.930 | 0.977 | 0.953 | 0.910 | 0.869 | 0.898 | 0.883 | 0.791 | 0.839 | 0.979 | 0.903 | 0.824 | 0.940 | 0.990 | 0.964 | 0.931 |
| 21 | 0.941 | 0.829 | 0.882 | 0.788 | 0.766 | 0.874 | 0.816 | 0.689 | 0.830 | 0.859 | 0.844 | 0.730 | 0.952 | 0.935 | 0.944 | 0.893 |
| 22 | 0.891 | 0.836 | 0.862 | 0.758 | 0.768 | 0.741 | 0.754 | 0.606 | 0.927 | 0.794 | 0.856 | 0.748 | 0.897 | 0.962 | 0.928 | 0.866 |
| 23 | 0.976 | 0.959 | 0.968 | 0.937 | 0.870 | 0.970 | 0.917 | 0.847 | 0.938 | 0.981 | 0.959 | 0.920 | 0.995 | 0.986 | 0.990 | 0.981 |
| 24 | 0.855 | 0.850 | 0.853 | 0.743 | 0.732 | 0.738 | 0.735 | 0.581 | 0.941 | 0.818 | 0.875 | 0.778 | 0.846 | 0.966 | 0.902 | 0.821 |
| 25 | 0.913 | 0.947 | 0.930 | 0.869 | 0.883 | 0.751 | 0.812 | 0.683 | 0.929 | 0.951 | 0.940 | 0.886 | 0.985 | 0.951 | 0.968 | 0.937 |
| 26 | 0.917 | 0.809 | 0.859 | 0.753 | 0.814 | 0.734 | 0.772 | 0.629 | 0.913 | 0.878 | 0.895 | 0.810 | 0.929 | 0.956 | 0.942 | 0.890 |
| 27 | 0.979 | 0.920 | 0.948 | 0.901 | 0.736 | 0.732 | 0.734 | 0.580 | 0.983 | 0.912 | 0.947 | 0.898 | 0.989 | 0.959 | 0.974 | 0.949 |
| 28 | 0.836 | 0.867 | 0.851 | 0.741 | 0.786 | 0.725 | 0.754 | 0.605 | 0.870 | 0.824 | 0.846 | 0.733 | 0.913 | 0.925 | 0.919 | 0.850 |
| 29 | 0.941 | 0.931 | 0.936 | 0.879 | 0.898 | 0.807 | 0.850 | 0.739 | 0.944 | 0.940 | 0.942 | 0.890 | 0.990 | 0.980 | 0.985 | 0.970 |
| 30 | 0.804 | 0.909 | 0.853 | 0.744 | 0.756 | 0.791 | 0.773 | 0.630 | 0.828 | 0.948 | 0.884 | 0.792 | 0.948 | 0.948 | 0.948 | 0.902 |
| 31 | 0.821 | 0.756 | 0.787 | 0.649 | 0.744 | 0.796 | 0.769 | 0.624 | 0.855 | 0.864 | 0.859 | 0.753 | 0.897 | 0.924 | 0.910 | 0.836 |
| 32 | 0.968 | 0.792 | 0.871 | 0.772 | 0.797 | 0.769 | 0.783 | 0.643 | 0.864 | 0.761 | 0.809 | 0.680 | 0.934 | 0.929 | 0.932 | 0.872 |
| 33 | 0.911 | 0.888 | 0.899 | 0.817 | 0.809 | 0.759 | 0.783 | 0.643 | 0.880 | 0.857 | 0.868 | 0.767 | 0.957 | 0.875 | 0.914 | 0.842 |
| 34 | 0.964 | 0.750 | 0.844 | 0.730 | 0.781 | 0.714 | 0.746 | 0.595 | 0.858 | 0.845 | 0.851 | 0.741 | 0.976 | 0.839 | 0.903 | 0.822 |
| 35 | 0.869 | 0.879 | 0.874 | 0.776 | 0.780 | 0.749 | 0.764 | 0.618 | 0.828 | 0.915 | 0.869 | 0.769 | 0.957 | 0.853 | 0.902 | 0.822 |
| 36 | 0.986 | 0.714 | 0.828 | 0.706 | 0.839 | 0.696 | 0.761 | 0.614 | 0.826 | 0.856 | 0.841 | 0.725 | 0.947 | 0.923 | 0.935 | 0.877 |
| 37 | 0.853 | 0.889 | 0.871 | 0.771 | 0.838 | 0.749 | 0.791 | 0.654 | 0.903 | 0.882 | 0.892 | 0.805 | 0.954 | 0.986 | 0.970 | 0.941 |
| Average | 0.912 | 0.904 | 0.906 | 0.831 | 0.827 | 0.824 | 0.825 | 0.708 | 0.899 | 0.897 | 0.897 | 0.816 | 0.953 | 0.954 | 0.953 | 0.911 |

| | MSF-Trans | | | | DC-Net | | | | ViT-Base | | | | ViT-Tiny (Ours) | | | |
|---|---|---|---|---|---|---|---|---|---|---|---|---|---|---|---|---|
| Class | Precision | Recall | F1-Score | Accuracy | Precision | Recall | F1-Score | Accuracy | Precision | Recall | F1-score | Accuracy | Precision | Recall | F1-score | Accuracy |
| 0 | 0.990 | 1.000 | 0.995 | 0.995 | 0.940 | 0.910 | 0.925 | 0.997 | 1.000 | 1.000 | 1.000 | 1.000 | 1.000 | 1.000 | 1.000 | 1.000 |
| 1 | 0.970 | 0.978 | 0.975 | 0.983 | 0.930 | 0.970 | 0.950 | 0.978 | 0.991 | 0.995 | 0.993 | 0.986 | 0.995 | 0.986 | 0.991 | 0.982 |
| 2 | 0.980 | 0.986 | 0.982 | 0.991 | 0.950 | 0.930 | 0.940 | 0.965 | 0.995 | 0.995 | 0.995 | 0.990 | 0.973 | 1.000 | 0.986 | 0.973 |
| 3 | 0.980 | 0.970 | 0.975 | 0.955 | 0.960 | 0.910 | 0.934 | 0.944 | 1.000 | 0.976 | 0.988 | 0.976 | 0.990 | 0.981 | 0.986 | 0.972 |
| 4 | 0.980 | 0.995 | 0.986 | 0.995 | 0.930 | 0.970 | 0.950 | 0.998 | 0.969 | 0.996 | 0.982 | 0.965 | 0.977 | 1.000 | 0.988 | 0.977 |
| 5 | 0.980 | 0.990 | 0.985 | 0.995 | 0.990 | 1.000 | 0.995 | 0.938 | 0.995 | 0.986 | 0.990 | 0.981 | 0.979 | 0.995 | 0.987 | 0.974 |
| 6 | 0.960 | 0.979 | 0.969 | 0.972 | 0.900 | 0.940 | 0.920 | 0.958 | 0.938 | 1.000 | 0.968 | 0.938 | 1.000 | 1.000 | 1.000 | 1.000 |
| 7 | 0.960 | 0.988 | 0.976 | 0.988 | 0.600 | 0.880 | 0.714 | 0.934 | 0.990 | 1.000 | 0.995 | 0.990 | 0.975 | 1.000 | 0.987 | 0.975 |
| 8 | 0.960 | 0.813 | 0.881 | 0.844 | 0.970 | 0.930 | 0.950 | 1.000 | 1.000 | 0.988 | 0.994 | 0.988 | 1.000 | 1.000 | 1.000 | 1.000 |
| 9 | 0.970 | 0.980 | 0.973 | 0.976 | 0.940 | 0.940 | 0.940 | 0.992 | 0.976 | 0.967 | 0.972 | 0.945 | 0.981 | 0.976 | 0.978 | 0.957 |
| 10 | 0.960 | 0.995 | 0.979 | 0.995 | 0.920 | 0.990 | 0.954 | 0.979 | 0.991 | 0.973 | 0.982 | 0.964 | 0.975 | 0.995 | 0.985 | 0.970 |
| 11 | 0.980 | 0.981 | 0.978 | 0.976 | 0.920 | 0.960 | 0.940 | 0.985 | 0.971 | 0.995 | 0.983 | 0.967 | 0.963 | 1.000 | 0.981 | 0.963 |
| 12 | 0.970 | 1.000 | 0.984 | 0.995 | 0.970 | 0.890 | 0.928 | 0.967 | 0.972 | 0.968 | 0.970 | 0.942 | 1.000 | 1.000 | 1.000 | 1.000 |
| 13 | 0.990 | 0.968 | 0.978 | 0.979 | 0.960 | 0.920 | 0.940 | 0.993 | 0.995 | 0.970 | 0.982 | 0.965 | 0.990 | 0.960 | 0.975 | 0.951 |
| 14 | 0.960 | 0.986 | 0.975 | 0.977 | 0.910 | 0.980 | 0.944 | 0.961 | 0.994 | 0.989 | 0.992 | 0.984 | 0.990 | 0.976 | 0.983 | 0.967 |
| 15 | 0.970 | 0.957 | 0.962 | 0.957 | 0.940 | 0.970 | 0.955 | 0.983 | 0.975 | 0.995 | 0.985 | 0.970 | 0.985 | 0.961 | 0.973 | 0.947 |
| 16 | 0.980 | 0.980 | 0.980 | 0.969 | 0.960 | 0.940 | 0.950 | 0.928 | 0.978 | 1.000 | 0.989 | 0.978 | 0.985 | 0.995 | 0.990 | 0.981 |
| 17 | 0.960 | 0.978 | 0.960 | 0.969 | 0.980 | 0.890 | 0.933 | 0.939 | 0.996 | 1.000 | 0.998 | 0.996 | 0.989 | 0.978 | 0.984 | 0.968 |
| 18 | 1.000 | 0.977 | 0.986 | 0.986 | 0.940 | 0.910 | 0.925 | 0.923 | 1.000 | 0.985 | 0.992 | 0.985 | 0.977 | 0.982 | 0.980 | 0.960 |
| 19 | 0.957 | 0.978 | 0.967 | 0.978 | 0.950 | 0.910 | 0.930 | 0.946 | 0.971 | 0.976 | 0.973 | 0.948 | 0.989 | 0.989 | 0.989 | 0.979 |
| 20 | 0.995 | 0.995 | 0.995 | 1.000 | 0.960 | 0.920 | 0.940 | 0.907 | 0.981 | 0.990 | 0.986 | 0.971 | 0.987 | 0.991 | 0.989 | 0.978 |
| 21 | 0.975 | 0.970 | 0.972 | 0.965 | 0.980 | 0.880 | 0.927 | 0.903 | 0.976 | 0.990 | 0.983 | 0.966 | 0.981 | 0.976 | 0.978 | 0.958 |
| 22 | 0.990 | 0.961 | 0.975 | 0.951 | 0.990 | 0.960 | 0.975 | 0.889 | 0.957 | 0.980 | 0.968 | 0.939 | 0.995 | 0.946 | 0.970 | 0.941 |
| 23 | 0.992 | 0.990 | 0.991 | 0.992 | 0.920 | 1.000 | 0.958 | 0.894 | 0.995 | 0.977 | 0.986 | 0.973 | 0.997 | 1.000 | 0.999 | 0.997 |
| 24 | 0.928 | 0.973 | 0.950 | 0.968 | 0.930 | 0.910 | 0.920 | 0.914 | 0.949 | 0.989 | 0.969 | 0.939 | 0.971 | 0.990 | 0.980 | 0.961 |
| 25 | 0.978 | 0.994 | 0.986 | 0.983 | 0.970 | 0.970 | 0.970 | 0.925 | 0.983 | 1.000 | 0.992 | 0.983 | 0.990 | 1.000 | 0.995 | 0.990 |
| 26 | 0.985 | 0.956 | 0.970 | 0.971 | 0.970 | 0.930 | 0.950 | 0.905 | 0.947 | 0.936 | 0.941 | 0.889 | 0.989 | 0.989 | 0.989 | 0.978 |
| 27 | 0.949 | 0.959 | 0.954 | 0.964 | 0.950 | 0.910 | 0.930 | 0.883 | 0.972 | 0.989 | 0.980 | 0.962 | 0.995 | 0.991 | 0.993 | 0.986 |
| 28 | 0.967 | 0.953 | 0.960 | 0.939 | 0.980 | 0.970 | 0.975 | 0.905 | 0.981 | 0.976 | 0.978 | 0.957 | 0.985 | 0.969 | 0.977 | 0.955 |
| 29 | 0.996 | 0.965 | 0.980 | 0.969 | 0.890 | 1.000 | 0.942 | 0.923 | 0.953 | 1.000 | 0.976 | 0.953 | 0.981 | 0.995 | 0.988 | 0.977 |
| 30 | 0.925 | 0.970 | 0.947 | 0.965 | 0.900 | 0.940 | 0.920 | 0.915 | 0.971 | 1.000 | 0.985 | 0.971 | 0.950 | 1.000 | 0.975 | 0.950 |
| 31 | 0.963 | 0.953 | 0.958 | 0.963 | 0.990 | 0.880 | 0.932 | 0.883 | 0.985 | 0.975 | 0.980 | 0.961 | 0.958 | 0.976 | 0.967 | 0.936 |
| 32 | 0.989 | 0.939 | 0.963 | 0.933 | 0.970 | 0.930 | 0.950 | 0.862 | 1.000 | 0.972 | 0.986 | 0.972 | 0.995 | 0.940 | 0.967 | 0.935 |
| 33 | 0.964 | 0.954 | 0.959 | 0.974 | 0.980 | 0.940 | 0.960 | 0.890 | 0.966 | 0.970 | 0.968 | 0.938 | 0.982 | 0.982 | 0.982 | 0.964 |
| 34 | 0.953 | 0.951 | 0.952 | 0.944 | 0.960 | 0.970 | 0.965 | 0.870 | 0.957 | 0.952 | 0.955 | 0.913 | 0.970 | 0.961 | 0.966 | 0.934 |
| 35 | 0.970 | 0.918 | 0.943 | 0.923 | 0.990 | 0.960 | 0.975 | 0.863 | 0.995 | 0.971 | 0.983 | 0.966 | 0.975 | 0.970 | 0.973 | 0.947 |
| 36 | 0.956 | 0.921 | 0.938 | 0.926 | 0.950 | 0.890 | 0.919 | 0.864 | 1.000 | 0.948 | 0.973 | 0.948 | 0.994 | 0.968 | 0.981 | 0.962 |
| 37 | 0.957 | 0.955 | 0.956 | 0.970 | 0.920 | 0.920 | 0.920 | 0.882 | 0.986 | 0.954 | 0.969 | 0.941 | 0.990 | 0.985 | 0.988 | 0.976 |
| Average | 0.970 | 0.967 | 0.968 | 0.968 | 0.941 | 0.937 | 0.938 | 0.931 | 0.980 | 0.982 | 0.981 | 0.963 | **0.984** | **0.984** | **0.984** | **0.969** |

V. THREATS TO VALIDITY

*A. Internal Validity*

This work has one potential threat to internal validity is the reliance on the WM-38k dataset, which may not fully represent all possible semiconductor defect variations. While data augmentation helps mitigate this, biases in the dataset could influence the model's generalization. Another concern is hyperparameter selection, as fine-tuning ViT-Tiny on specific configurations may lead to performance variations across different settings. Without extensive cross-validation or additional independent test sets, the reported accuracy may not fully reflect real-world model performance.

*B. External Validity*

The generalizability of ViT-Tiny beyond the WM-38k dataset remains a concern, as real-world semiconductor defect distributions may differ. Additionally, the model's performance in unseen manufacturing environments with varying imaging conditions and noise levels is uncertain. Further testing on diverse datasets and industrial settings is needed to confirm its robustness.

VII. CONCLUSION

Semiconductor defect detection is a fundamental challenge in modern manufacturing, where precision, efficiency, and scalability are crucial to maintaining high production yield and quality assurance. In this study, we explored Large Vision Models (LVMs) for semiconductor defect classification and proposed an optimized ViT-Tiny-based framework that effectively addresses these limitations. Our approach leverages domain-specific fine-tuning of the pretrained ViT-Tiny model, which significantly reduces space and time complexity while maintaining overall high classification F1-Score 98.4%. Unlike larger ViT variants and other state-of-the-art (SOTA) methods, ViT-Tiny demonstrates a more computationally efficient architecture, making it a viable solution for industrial applications where real-time defect detection is critical. Through extensive experimentation, our framework achieves remarkable per-class classification accuracy, F1-score, precision and recall, outperforming existing techniques in recognizing both single-defect and mixed-defect types, particularly 2-, 3-, and 4-mixed defects. These results demonstrate the model's robustness in handling complex defect patterns, which is a key requirement in semiconductor manufacturing.

Future research directions include further optimizing the ViT-Tiny architecture for even lower computational overhead, investigating its applicability to other industrial defect detection tasks, and exploring multimodal learning techniques that incorporate additional sensor data and image description as an additional modality to enhance classification accuracy.